# Proximity Measure of Information Object Features for Solving the Problem of Their Identification in Information Systems

**V.V. Yuzefovych** Institute for Information Recording of the National Academy of Sciences of Ukraine, 2 Mykoly Shpaka St, Kyiv, Ukraine.

**Abstract**

The paper considers a new quantitative-qualitative proximity measure for the features of information objects, where data enters a common information resource from several sources independently. The goal is to determine the possibility of their relation to the same physical object (observation object). The proposed measure accounts for the possibility of differences in individual feature values - both quantitative and qualitative - caused by existing determination errors. To analyze the proximity of quantitative feature values, the author employs a probabilistic measure; for qualitative features, a measure of possibility is used. The paper demonstrates the feasibility of the proposed measure by checking its compliance with the axioms required of any measure. Unlike many known measures, the proposed approach does not require feature value transformation to ensure comparability. The work also proposes several variants of measures to determine the proximity of information objects (IO) based on a group of diverse features.

---

**Introduction**

Any information system tasked with collecting and processing information about environmental objects may encounter situations where data of varying quality regarding the same objects arrive separately from multiple internal sources (or external systems) as data concerning different objects. This phenomenon arises because different information sources, even when solving tasks in coordination, effectively function independently. For objective reasons, they are often unable to determine which objects they share. Consequently, absent special measures or additional information processing within the system, information objects relating to the same real environmental objects may (and evidently will) exist in parallel within the information resource.

Here, an **Information Object (IO)** is understood as a combined set of feature values of a really existing **Physical Object (PO)**, available for observation and determined by the system's data collection means. In such cases, a task arises to identify IO instances (attributing them to common POs) with the subsequent unification of IO characteristics. Solving this identification task allows one to eliminate information duplication in the system.

Duplication leads not only to an unproductive increase in stored data volumes but also to an erroneous assessment of the object saturation of the analyzed environment. Furthermore, combining IO data from multiple sources improves the quality (completeness, accuracy, reliability) of the received information, thereby reducing information uncertainty regarding the objects of interest.

The task of identifying IO instances can be solved quite effectively by analyzing the behavioral characteristics of observation objects and the capabilities of the data acquisition subsystem (environmental monitoring subsystem). In particular, relevant approaches are widely used in surveillance systems for air, ground, or surface objects through the application of so-called tertiary information processing [1], which involves merging data from many sources regarding the same objects.

Another, generally less effective, approach to IO identification (outside the data acquisition subsystem) is the analysis of the similarity of characteristics (feature values) of information objects arriving from different sources. Reference [2] describes such an approach to IO identification in the information space based on a set of features. It proposes a step-by-step analysis of IO characteristics, assuming that two different objects with absolutely identical characteristics cannot exist in the information space. That is, an object is considered identified if all its features fully coincide with the features of another IO already existing in the system. Features may include not only intrinsic characteristics but also interrelations with other information objects.

It should be noted that the approach proposed in [2] cannot be considered universal, as it assumes a complete match of individual feature values (admitting only the absence of data for a specific feature). In reality, any information source determines object characteristics (feature values) with a certain precision (error). Consequently, a complete coincidence of IO characteristics obtained from different sources regarding the same object will not occur in the general case.

Thus, the problem arises from defining a proximity (similarity) measure. Based on the analysis of this measure, one could decide that "similar" information objects might be identified as the same physical object, taking into account the errors in feature value determination. Depending on the context and usage in primary sources, this paper will use the phrases "proximity measure" (sometimes identical to "similarity measure") or "distance measure", as an inverse relationship exists between them.

**Problem Statement**

As is known, for objects described by quantitative indicators, common distance measures include Linear (Manhattan) distance, Euclidean distance, Generalized Minkowski distance, Mahalanobis distance, and others [3, 4]. These require the prior normalization of features measured in different units (metric scales) to ensure formal comparability [3].

For qualitative features, proximity (similarity) measures such as Rao, Hamming, Rogers-Tanimoto, and Jaccard coefficients are used. These involve analyzing (essentially counting) features characterized by a complete match or mismatch of their values [4, 5].

However, in most cases, IOs are described simultaneously by quantitative and qualitative features. Therefore, the proximity (distance) measure between IOs must be quantitative-qualitative.

● **Quantitative features:** Features subject to measurement by various technical means or calculated based on such measurements.

● **Qualitative features:** Features determined through active human reasoning. Note that qualitative features may also be denoted by numbers, but they do not lose their qualitative (subjective) character.

Universal quantitative-qualitative measures include Zhuravlev's proximity measure, Voronin's approximation proximity measure, Mirkin's similarity measure, and others. Among these, the meaning of Zhuravlev's measure is considered the most "transparent" [3]:

$$\rho_{ij} = \sum_{l=1}^{L} \alpha_{ij}^{l},$$

where $l$ - the object feature index ($l = \overline{1, L}$);

$$\alpha_{ij}^{l} \text{ (for quantitative features)} = \begin{cases} 1, \text{ if } \left| x_i^l - x_j^l \right| \leq \varepsilon^l \\ 0, \text{ in other case} \end{cases} ;$$

$$\alpha_{ij}^{l} \text{ (for qualitative features)} = \begin{cases} 1, & \text{if the feature is exist} \\ 0, & \text{in other case} \end{cases}$$

$\varepsilon^{l}$ - the quantitative proximity threshold for the $l$-th feature.

As seen from the expression, Zhuravlev's measure accounts for the possibility of differences in quantitative object features (not exceeding a certain value $\varepsilon^{l}$), under which it is accepted that the values actually coincide. However, the measure does not envisage such a possibility for qualitative features. They must coincide completely; otherwise, the values are considered different. Yet, in reality, qualitative feature values (e.g., those determined on ordinal scales) can also be close in meaning and lie within a certain error margin. Effectively, Zhuravlev's measure reduces all features to binary qualitative ones and does not account for a gradual (smooth) change in proximity proportional to value differences.

Reference [6] proposes a combined similarity degree coefficient constructed by determining similarity for each feature type separately:

$$\rho_{ij} = \frac{m_K \delta_{ij} + m_Q h_{ij}}{m_K + m_Q}$$

where $m_K$, $m_Q$ – the number of quantitative and qualitative features, respectively;

$\delta_{ij}$ – any known quantitative proximity measure (meeting the requirements for measures) between objects $i$ and $j$ (normalized to [0, 1]);

$h_{ij}$ – the Hamming coefficient normalized by the number of qualitative features.

If $m_K$ and $m_Q$ are treated as weights, this corresponds to a weighted average. This approach also requires normalization of quantitative indicators, while qualitative indicators again rely on complete matches or mismatches (Hamming coefficient).

**The aim of this article** is to propose a new quantitative-qualitative distance measure between IO features to solve the identification task. This measure avoids the aforementioned drawbacks, specifically by accounting for errors in determining both quantitative and qualitative feature values. To solve this, we first consider quantitative and qualitative features separately.

**Determination of Proximity (Distance) Measure for Quantitative Features**

As is known, the measurement error of any quantitative feature by a technical device can be described by a probability distribution law. Figure 1 illustrates several measurement results of a conditional magnitude $v$ ($v_1$, $v_2$, $v_3$, $v_4$) by different devices, where the true unknown value is $v_0$ in all cases. The measurement errors follow identical laws but with different parameters.

Let the measurement error equal the value corresponding to the Root Mean Square Error (RMSE or σ) for each device: $\sigma_1$, $\sigma_2$, $\sigma_3$, $\sigma_4$. Obviously, differences in σ values mean that linear distances (like Euclidean or Minkowski) between measured values (e.g., $r_{12}$ and $r_{34}$) will differ. However, the probability that the actual feature value lies within one RMSE of each device is identical in all cases. It is also evident that the probability that obtained values relate to the same actual value increases non-linearly as the linear distance $r_{ij}$ decreases (given fixed RMSE), according to error distribution laws. Furthermore, a mere coincidence of measured values (obtained with errors) does not guarantee a coincidence of actual values.

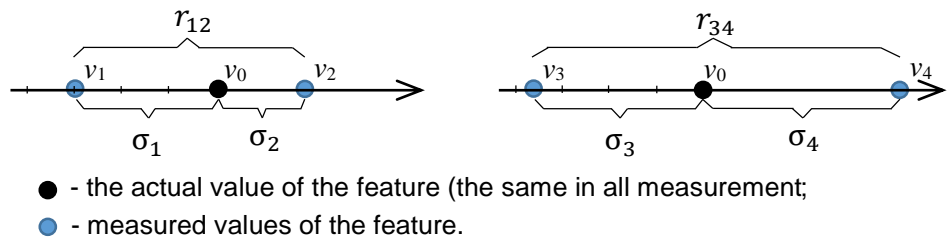

● - the actual value of the feature (the same in all measurement;
● - measured values of the feature.

**Figure 1** – *Results of measuring a certain quantity by different measuring devices.*

Thus, the probability that the actual feature value is the same for both measurements serves as a robust distance (proximity) measure.

Since determining error distribution laws for every device is tedious, one can assume a normal distribution (justified by the Central Limit Theorem). To determine the distance between quantitative values, one needs the values, the RMSE ($\sigma$) of the devices, and the mathematical expectations ($m$). If the RMSE is unknown, it can be estimated from the maximum absolute measurement error ($\Delta_{max}$), typically specified for instruments. Using the "three-sigma" rule (ignoring the negligible probability of errors exceeding $3\sigma$), the RMSE is:

$$\sigma \approx \Delta_{max}/3.$$

More complex systems provide statistical accuracy indicators, usually including the RMSE. However, the mathematical expectations of errors are unknown here. To solve this, we propose estimating the possibility that measured values are potentially the same by determining the probability of finding the actual value within a common range.

We use the probability calculation for a normal distribution in an interval ($c$, $d$):

$$P(c \leq x \leq d) = \Phi\left(\frac{d-m}{\sigma}\right) - \Phi\left(\frac{c-m}{\sigma}\right),$$

where $\Phi(\cdot)$ – the Laplace function.

Given independent measurements, the probability that the quantity lies in a specific range is the product $P_S = P_{i(c,d)} \cdot P_{j(c,d)}$[1]. For calculation, we treat the values from sources $i$ and $j$ as the most probable values (mathematical expectations)[2].

Figure 2 shows an example: the results of measuring feature $X$ by two measuring devices (measurement results $x_1=12$ and $x_2=18$) and the Gaussian error distribution laws for each measurement, obtained based on the $\sigma$ data for each measurement ($\sigma_{x_1}=3$, $\sigma_{x_2}=2$). The intersection of probable value ranges (based on the three-sigma rule) is defined by the interval [$c$, $d$], where: $c = max(x_1 - 3\sigma_1, x_2 - 3\sigma_2)$, $d = min(x_1 + 3\sigma_1, x_2 + 3\sigma_2)$.

For the example in Figure 2: Interval 1: [12-9, 12+9] = [3, 21]. Interval 2: [18-6, 18+6] = [12, 24]. Intersection: [12, 21].

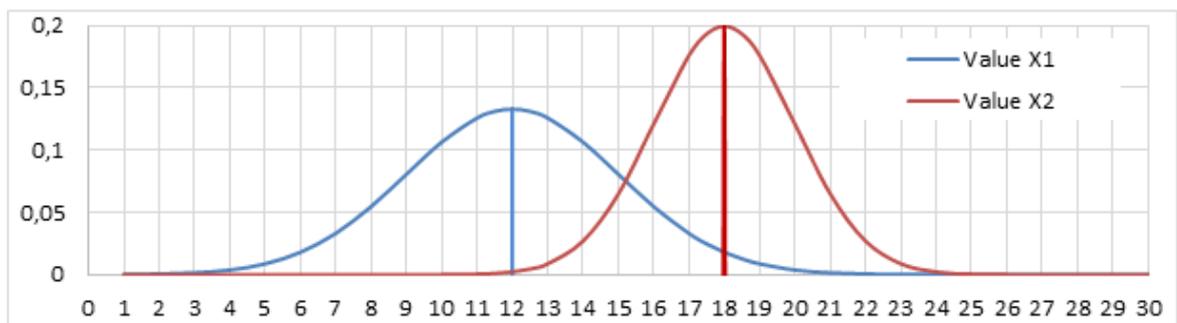

**Figure 2** – *Error distribution laws for values measured by two different sources.*

Next, we calculate the probability that the true value for each measurement lies within [12, 21]. For example:

---

[1] In our case, the errors from different sources are uncorrelated. If there is a correlation between the errors, the proposed solution will require modification, which is beyond the scope of this work.

[2] In fact, the most probable value of the attribute should be its true value (in the absence of systematic measurement error), which is unknown in the problem. However, given that this assumption applies to all quantitative attributes of all IOs, the nature of the distortion of reality will not lead to a violation of the adequacy of the measure, which will be shown below.

$$P_{x_1}(12 \le x \le 21) = \Phi\left(\frac{21-12}{3}\right) - \Phi\left(\frac{12-12}{3}\right) = 0{,}499,$$

$$P_{x_2}(12 \le x \le 21) = \Phi\left(\frac{21-18}{2}\right) - \Phi\left(\frac{12-18}{2}\right) = 0{,}932.$$

The joint probability is $P_{S x_i x_j} = 0{,}499 \cdot 0{,}932 \approx 0{,}465$. Thus, the proximity measure for one quantitative feature is $\rho'_{K_{ij}} = P_{S x_i x_j}$.

To obtain a distance measure ($\rho_K$), which increases with difference, we invert the probability:

$$\rho_{K_{ij}} = 1 - \rho'_{K_{ij}}.$$

If distributions and values coincide completely, $P_{S x_i x_j} \approx 1$ (proximity measure $\rho'_{K_{ij}}$) and the distance $\rho_{K_{ij}} \approx 0$.

An obvious advantage of using a probabilistic measure to calculate the distance (proximity) measure based on a single feature is that it ranges from 0 to 1, that is, it is inherently normalized. However, it is advisable to verify whether the obtained distance measure satisfies the axioms required for any measure.

As is known, a distance measure can be a certain quantity $d_{ij}$, that satisfies the following conditions [3]:

1. **Non-negativity:** $\forall (x_1, x_2): d(x_1, x_2) \ge 0$.
2. **Symmetry:** $\forall (x_1, x_2): d(x_1, x_2) = d(x_2, x_1)$.
3. **Identity:** $d(x_1, x_2) = 0,$ if $x_1 = x_2$.
4. **Triangle Inequality:** $\forall (x_1, x_2, x_3): d(x_1, x_2) \le d(x_1, x_3) + d(x_3, x_2)$.

**Verification against Axioms.** The fulfillment of the first three specified conditions follows from the very concept of probability and the rules for its calculation. For example, Figure 3 shows the dependence of the distance measure calculated for two IOs based on a single quantitative indicator, the data for which were determined with identical error $\sigma_1 = \sigma_2 = 2$.

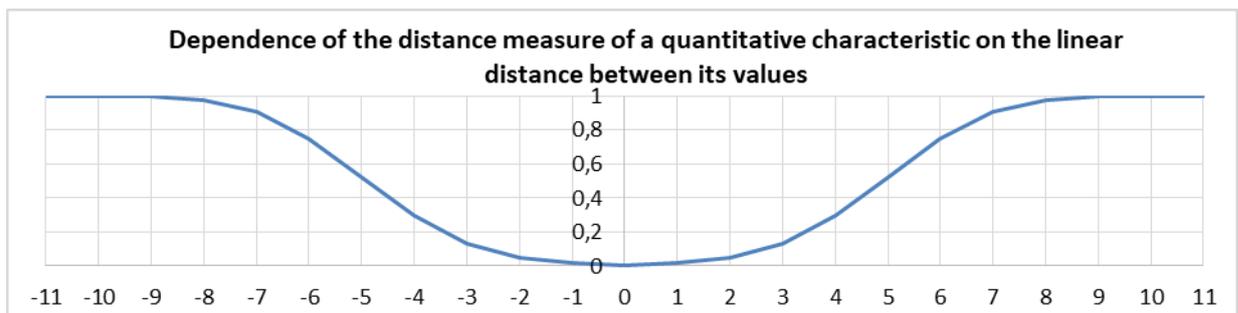

**Figure 3** – *Dependence of quantitative distance measure on linear distance between two values with errors.*

As can be seen from Figure 3, the proposed measure satisfies the conditions of non-negativity, symmetry, and identity (maximum similarity of the object with itself). Let us verify the fulfillment of the triangle inequality condition using an example. Let us assume we have three values of a single feature from three sources of equal $(\sigma_1 = \sigma_2 = \sigma_3 = 2)$ precision: $x_1=12$, $x_2=15$, $x_3=18$. Let us calculate three values of the distance measure using the expressions provided:

$$\rho(x_1, x_2) \approx 0{,}13; \; \rho(x_1, x_3) \approx 0{,}75; \; \rho(x_3, x_2) \approx 0{,}13.$$

Thus, the triangle inequality (as this case demonstrates) will not always be fulfilled, since, in particular $\rho(x_1, x_3) > \rho(x_1, x_2) + \rho(x_2, x_3)$. This is explained by the significant non-linearity in the growth of the probability of finding the feature value within a given interval as the linear distance between the obtained measurements decreases. At the same time, it is noted in [3] that the latter condition (the triangle inequality) is not always constructive, and in principle, distance measures that do not satisfy this inequality may be used. Thus, the proposed measure for quantitative features can be considered viable (as will be confirmed below), especially since it is based on probability analysis, which possesses concrete physical meaning.

Since the proposed measure claims to account for errors in determining feature values, let us analyze how the values of such errors affect the distance measure. Figure 4 shows the change in the distance value between feature values for the previous case shown in Figure 3; however, the dependence of this distance for $\sigma_1 = \sigma_2 = 1$ is added as a dashed line. As can be seen from the figure, the distance between feature values for the case $\sigma_1 = \sigma_2 = 1$ is larger and increases more rapidly in all instances (except for a complete coincidence of values), because smaller feature determination errors increase confidence in the obtained measurement values, and consequently, in the fact of their difference.

At the same time, the identical distance value for both cases when feature values coincide may be a certain drawback of the proposed measure. Indeed, if two more precise information sources determine a feature value to be identical, confidence in the result should be higher (and thus the distance smaller) than in the same situation for two less precise sources. To eliminate this drawback (to achieve the desired effect), it is advisable to multiply the obtained distance measure by a certain constant, the value of which should be inversely dependent on σ. In this case, the constant will not affect the nature of the change in the measure. Considering that smaller σ values correspond to a higher probability density in the region of the mathematical expectation of a random variable, a coefficient $P_{\xi_{ij}}$ can be used as such a constant, for example. This coefficient will be determined by the probability that the feature value lies within the limits of a certain fixed magnitude $\xi$ relative to the mathematical expectation of the feature, the value of which was received from sources $i$ and $j$.

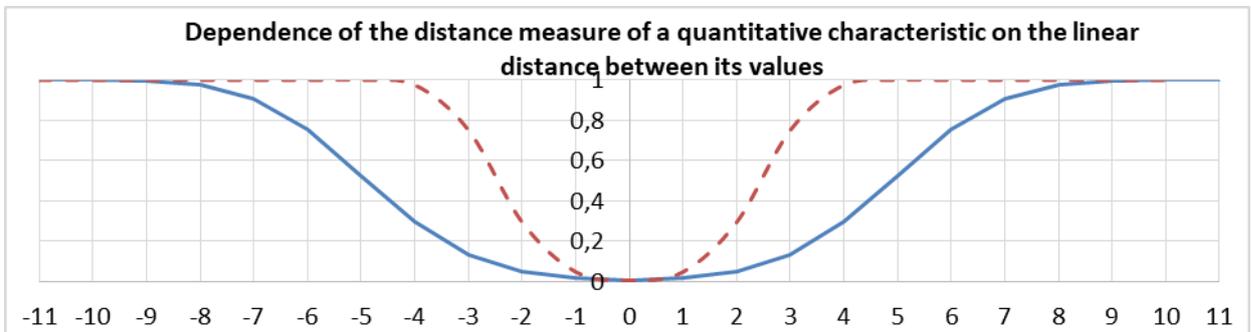

**Figure 4** – *Dependence of the quantitative feature distance measure for the case $\sigma_1 = \sigma_2 = 2$ (solid line) and the case $\sigma_1 = \sigma_2 = 1$ (dashed line) on the linear distance between two obtained feature values.*

As the magnitude $\xi$, one may choose the magnitude $3\sigma_{\min}$, where $\sigma_{\min}$ is the minimum root-mean-square error of feature determination among all available information sources, or of a conditional information source whose precision would be sufficient to establish confidence in the data obtained from it. Such a $\xi$ value will ensure a coefficient value $P_\xi \approx 1$ that preserves the range of change of the obtained measure within the limits of 0 to 1. For the case cited above, according to data on the more precise source, this magnitude will be: $\xi = 3$. Then, the probability that the random variable will be located within the limits $X_0 - \xi < x < X_0 + \xi$ will be:

- for case $\sigma_1 = 2, \sigma_2 = 2$: $P_{\xi_{12}} \approx 0{,}87$;
- for case $\sigma_1 = 1, \sigma_2 = 1$: $P_{\xi_{12}} \approx 1$;

- for case $\sigma_1 = 1, \sigma_2 = 2$: $P_{\xi_{12}} \approx 0{,}93$.

The value of the coefficient $P_{\xi_{ij}}$ can be obtained in any other way, provided the conditions put forward above are preserved.

Thus, finally, the proximity measure between quantitative feature values obtained from data of two different sources with different (but known) errors will be calculated using the expression:

$$\rho'_{K_{ij}} = P_{P_{S_{x_i x_j}}} \cdot P_{\xi_{ij}}$$

And, accordingly, for the distance measure:

$$\rho_{K_{ij}} = 1 - \rho'_{K_{ij}}.$$

Let us consider the determination of the proximity (distance) measure for qualitative IO features.

**Determination of the Proximity (Distance) Measure for Qualitative IO Features**

Let us analyze the task of calculating the possibility that two determined values of a specific qualitative feature are, in fact, a single value. Here, as previously noted, it is important to recall that a qualitative feature may be expressed numerically whilst retaining its qualitative character, since the "qualitative nature" of a feature is determined not by its form of expression (representation), but by the method of its derivation.

A significant aspect is the determination of the scale type on which the qualitative feature is defined. Generally, among qualitative scales, two substantially different scales are distinguished: nominal and ordinal [7]. Nominal scales are those where any value serves exclusively as a category name. Examples include department numbers within an organization, nationality, and car making. It is impossible to apply any arithmetic operations to the values of such scales, as they (the values) are not linked by any relationships. An ordinal scale represents a sequence of feature values, where each subsequent value characterizes the next rank within the set of possible values. The values of such a scale may be arranged according to a "greater-lesser" relationship. Examples of features determined on an ordinal scale include earthquake intensity, a person's professional level, and weather hazard level. A certain (albeit imprecise) distance may always be defined between such values; consequently, a specific number may be assigned to correspond to each feature value, upon which, in turn, arithmetic operations may be performed. Effectively, the transformation of an ordinal scale into a metric one, possessing units of "measurement", is possible.

Qualitative feature values obtained through direct human participation are characterized by non-statistical uncertainty. Currently, the application of Possibility Theory methods [8, 9] is considered most effective for the formalized analysis of such uncertainty. Thus, for ordinal scales, one may employ an approach to determine the distance measure between two given feature values based on their formalization as fuzzy sets (by assigning a corresponding membership function value to each possible feature value). The simplest for calculation purposes is a triangular membership function, the vertex of which corresponds to the most possible feature value. Figure 5 illustrates an example of obtaining two values of a certain feature $g$ as numbers ($G_1=12$ and $G_2=18$ units) from two different sources and formalizing them as two fuzzy sets with triangular membership functions, calculated using the expression:

$$\mu_G(g) = \mu_{max} \times \begin{cases} 0, & \text{if } g \leq g_{min}; \\ (g - g_{min})/(G - g_{min}), & \text{if } g_{min} < g < G; \\ 1 & \text{if } g = G; \\ (g_{max} - g)/(g_{max} - G), & \text{if } G < g < g_{max}; \\ 0, & \text{if } g \geq g_{max}, \end{cases}$$

where $g_{min}$, $g_{max}$ are the possible minimum and maximum feature values.

The maximum and minimum feature values may be calculated using the expressions:
$$g_{min} = \text{ROUND}(G - G \cdot k), \text{ or } \text{ROUND}(G \cdot (1 - k));$$
$$g_{max} = \text{ROUND}(G + G \cdot k), \text{ or } \text{ROUND}(G \cdot (1 + k)),$$
where ROUND() is the rounding function, and $k$ is a coefficient, the value of which is established based on the perception of the possible error in feature determination by a specific source according to the rule: the greater the possible error, the larger the number $k$ (for example, for $k=0,3$: $g_{min} = 0,7g$, a $g_{max} = 1,3g$).

The analogue of the multiplication operation (used when determining the probability of finding a feature from two sources within a defined value range) in fuzzy set theory is the intersection operation. The membership function of the resulting fuzzy set obtained from the intersection of two fuzzy sets constructed for $G_1$ and $G_2$ is:
$$\mu_{G_1 \wedge G_2}(x) = \min[\mu_{G_1}(x), \mu_{G_2}(x)].$$
Then, the possibility that both values in the messages are identical (MS) may be determined by the expression:
$$M_{S\,G_1,G_2} = \max\{\mu_{G_1 \wedge G_2}(x)\} = \max\{\min[\mu_{G_1}(x), \mu_{G_2}(x)]\}.$$

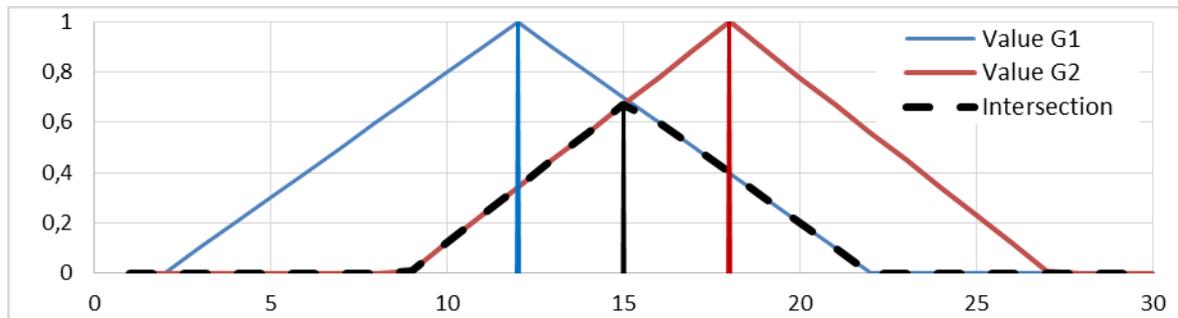

**Figure 5** – *Fuzzy sets describing possible feature values, provided by two sources, using triangular membership functions.*

For our example, $M_{S\,G_1,G_2} = 0,67$.

The obtained possibility can be considered as a proximity measure of the feature values:
$$\rho'_{Q_{ij}} = M_{S\,G_1,G_2}.$$
The Gaussian form of membership functions may also be used; however, the advisability of its application requires additional argumentation, if only because it leads to more complex calculations. Using analogous expressions, we obtain the result shown in Figure 6.

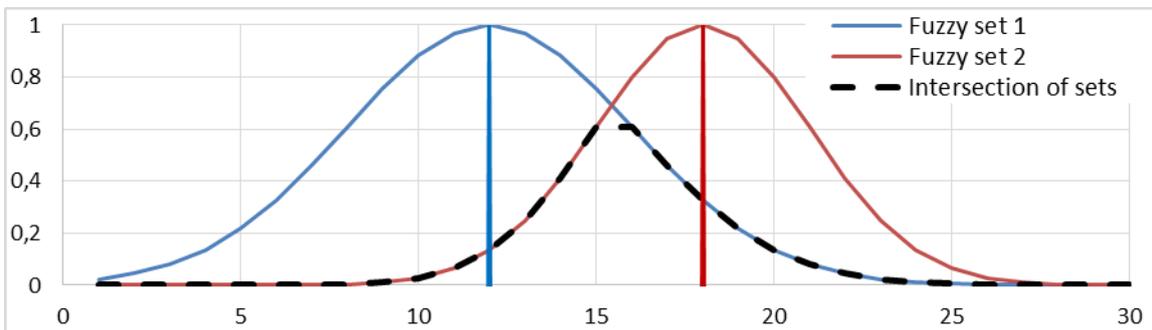

**Figure 6** - *Fuzzy sets describing possible parameter values using Gaussian membership functions (values for the intersection sets are calculated for integer values only).*

And the value $M_{S\,G_1,G_2} = \rho'_{Q_{ij}} = 0,61$ (corresponds to the feature value "15").

To obtain the distance measure, by analogy with quantitative features, it is also necessary to invert the obtained value, since as the obtained feature values approach one another, the value $M_{S_{ij}}$ must increase:

$$\rho_{Q_{ij}} = 1 - \rho'_{Q_{ij}}.$$

Evidently, the task is solved analogously for qualitative IO features defined by linguistic concepts on an ordinal scale. In this case, the corresponding fuzzy set is formed based on a term-set. Here, it is necessary to sort the feature values by the increase (strengthening) of the object property it characterizes and, as noted above, assign an ordinal number to each linguistic term.

If the feature is nominal, the membership function will be characterized by one extreme value and a certain identical permissible value for all other set values - $\Delta \in [0; 0,5)$[3], which will characterize the possibility of erroneous feature determination (Figure 7). Thus, if IO feature values arriving from two sources do not coincide, the distance (proximity) between them for this feature will be determined specifically by this value - $\Delta$, regardless of the feature values themselves.

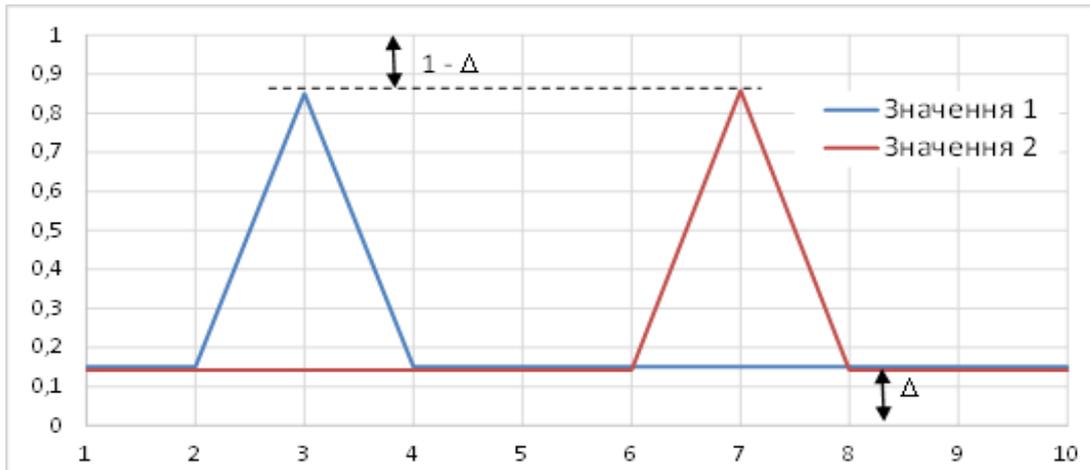

**Figure 7** – *Result of formalizing a nominal feature as a fuzzy set to determine the distance between its two values.*

It should be noted that, unlike measurement results, qualitative features are often additionally characterized by a degree of certainty (confidence) regarding their values. Evidently, a reduction in the certainty level of determined feature values must increase the distance (reduce the proximity) between such values, as the possibility that the features do not actually coincide increases. To account for the degree of certainty of the obtained value, one may introduce several for example, four certainty levels: 'Certain', 'Probable', 'Possible', 'Doubtful' [4], with corresponding numerical values. This ensures the possibility of formalizing this characteristic and performing algebraic calculations (for example, as shown in Table 1).

**Table 1** – *Numerical correspondence (scale) to defined linguistic certainty values*

| Linguistic certainty value | Numerical certainty value, $D_p$ |
|---|---|
| "Certain" determined feature value | 1 |
| "Probable" feature value is exactly this | 0,7 |
| "Possible" feature value is exactly this | 0,5 |
| "Doubtful" that feature value is exactly this | 0,25 |

If the obtained feature value possesses a certainty attribute lower than "certain", the membership function must be modified, accounting for the data in Table 1, in accordance with the expression:

$$\mu_{D_P}(g) = D_P \cdot \mu(g).$$

---

[3] A value of $\Delta = 0.5$ will result in loss of ability to identify the feature. A value of $\Delta > 0.5$ is meaningless.

[4] The number of levels (degrees) of certainty may also be different.

For example, if a specific feature value equals "1000" and possesses the certainty characteristic "probable", then, according to Table 1 and considering the latter expression, the membership function will appear as shown in Figure 8.

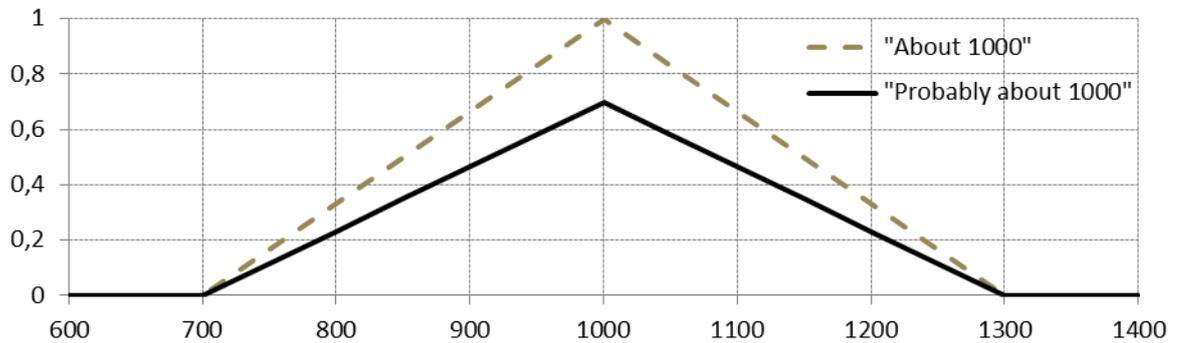

**Figure 8** – *Result of fuzzification of a numerical feature value: "probably about 1000".*

Let us verify the compliance of the proposed proximity measure for qualitative features with the conditions required of measures. As in the case of the analysis of the distance measure for quantitative features proposed above, it is evident that the obtained measure for qualitative features satisfies the aforementioned conditions of non-negativity, symmetry, and maximum similarity of an object with itself. Let us verify the fulfillment of the last condition – the triangle inequality. Figure 9 shows the results of estimating a certain qualitative feature value, determined on an ordinal scale, based on data from three sources, formalized as fuzzy sets.

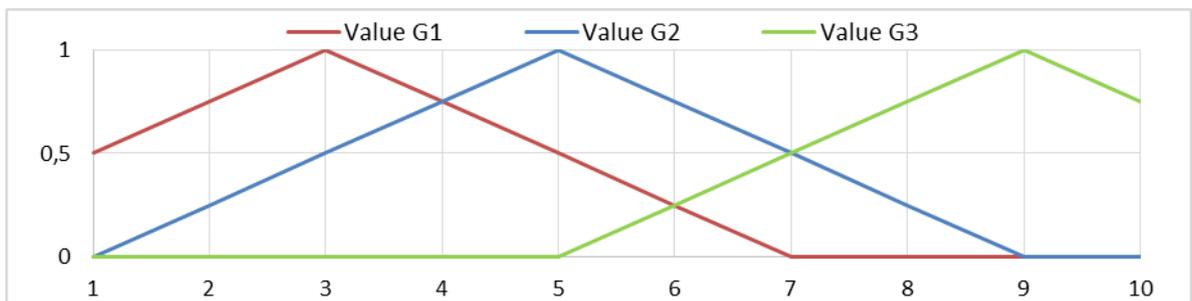

**Figure 9** – *Result of formalization of a nominal feature as a fuzzy set to determine the distance between two of its values.*

For this case: $d(G_1,G_2) = 1 - 0{,}75 = 0{,}25$; $d(G_1,G_3) = 1 - 0{,}25 = 0{,}75$; $d(G_3,G_2) = 1 - 0{,}5 = 0{,}5$ – thus, the triangle inequality holds. Let us consider the same case, but where one of the values is characterized by a degree of certainty lower than certain – "probable" (Figure 10).

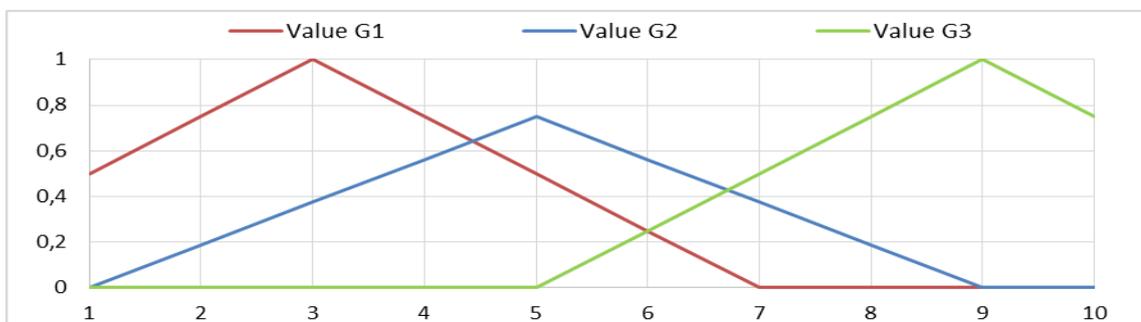

**Figure 10** – *Result of formalization of a nominal feature as a fuzzy set to determine the distance between two of its values.*

For this case: $d(G_1,G_2) = 1 - 0{,}56 \approx 0{,}44$; $d(G_1,G_3) = 1 - 0{,}25 \approx 0{,}75$; $d(G_3,G_2) = 1 - 0{,}375 \approx 0{,}63$. It is evident that the triangle inequality also holds.

Analysis of other feature value variants demonstrated that for the selected method of distance determination, the triangle inequality must always hold, since the change in distance between feature values in such a case possesses a linear character. We add that the fulfillment of the triangle inequality regarding quantitative features can also be ensured, if necessary, by "converting" the Gaussian form of error distribution laws into a triangular one through approximation.

Thus, by analogy with existing measures (specifically Zhuravlev's measure), for a generalized quantitative-qualitative distance measure characterizing the distance between two IOs, in general (considering the normalization of all values), one may write:

$$\rho_{Y_{ij}} = \sum_{l=1}^{L_1} \rho_{K_{ij}}^l + \sum_{l=L_1+1}^{L} \rho_{Q_{ij}}^l ,$$

where $L_1$ - the number of quantitative features;

$L$ - the total number of features.

Or, by analogy with the Rao coefficient (normalizing the quantitative-qualitative measure by the number of features), we write:

$$\rho_{Y_{ij}} = \sum_{l=1}^{L_1} \frac{\rho_{K_{ij}}^l}{L_1} + \sum_{l=L_1+1}^{L} \frac{\rho_{Q_{ij}}^l}{L - L_1} .$$

(Using separate feature counts $L_1$ and $L-L_1$ for normalization instead of a single value $L$ is advisable when the number of qualitative and quantitative IO features differs significantly and there is a need to neutralize the increased influence of one feature type on the resulting distance). In other cases, the value $L$ may suffice. Such a distance measure ($\rho_{Y_{ij}}$), unlike other known ones, allows accounting for possible measurement errors of quantitative object features and errors in determining qualitative features. The smaller the measure value (the closer its value to 0), the greater the object similarity and, consequently, the possibility that the information descriptions of objects via the set of features relate to the same PO.

Considering the possible varying influence (different weight) of quantitative and qualitative information, an expression for a weighted measure may also be written:

$$\rho_{Y_{ij}} = \sum_{l=1}^{L_1} w^l \rho_{K_{ij}}^l + \sum_{l=L_1+1}^{L} w^l \rho_{Q_{ij}}^l ,$$

where $w^l$ is the weight of the $l$-th feature, varying from 0 to 1, such that $w^1 + w^2 + ... + w^l = 1$.

For a simpler case (equal weight of all quantitative and all qualitative features among themselves), one may write:

$$\rho_{Y_{ij}} = w \sum_{l=1}^{L_1} \rho_{K_{ij}}^l + (1-w) \sum_{l=L_1+1}^{L} \rho_{Q_{ij}}^l .$$

(where $w$ is the feature weight varying from 0 to 1), or:

$$\rho_{Y_{ij}} = w \sum_{l=1}^{L_1} \frac{\rho_{K_{ij}}^l}{L_1} + (1-w) \sum_{l=L_1+1}^{L} \frac{\rho_{Q_{ij}}^l}{L - L_1} .$$

An expression for the weighted average value may also be written by analogy with [6], or another variant of a general expression combining distances between quantitative and qualitative IO features may be formed.

The most acceptable expression variant for calculating the distance measure between IOs for each task and subject area should be selected individually; however, all expressions presented above for obtaining $\rho_{Y_{ij}}$ are additive. Consequently, a large distance between values of certain features may be compensated by small distances in others through accumulation. This is evidently unacceptable for solving the identification task, where obtaining a large distance (low similarity) even on a single feature (e.g., coordinates) suffices to form a conclusion regarding data sets relating to different objects. To solve the identification task, the use of multiplicative convolutions of similarity indicators for individual features appears adequate. The simplest multiplicative convolution assumes equal weight for all indicators, equal to unity. Then, for direct action indicators (in our case – proximity indicators), we have a simple multiplication of all their values. Consequently, the final (aggregate across the entire set of features) similarity value of information objects will be less than the smallest one. If the similarity between IOs for one feature equals zero, the total similarity will also equal zero. A "softer" multiplicative convolution (for proximity indicators) will have the known form:

$$\rho'_{Y_{ij}} = \prod_{l=1}^{L_1} \rho'^{lw^l}_{K_{ij}} \cdot \prod_{l=L_1+1}^{L} \rho'^{lw^l}_{Q_{ij}}, \text{ where } \sum_{l=1}^{L} w^l = 1.$$

Accordingly, for the distance measure:

$$\rho_{Y_{ij}} = 1 - \rho'_{Y_{ij}}.$$

Let us conduct a modeling experiment regarding the calculation of the quantitative-qualitative proximity measure between IOs in the form of a multiplicative convolution of proximity measure values for individual features, using the expression provided. IO data (planar rectangular coordinates and type) arrive from two sources of differing precision: Coordinate determination RMSE of 20 and 30 meters (the best RMSE among all possible sources is 10m); IO type is determined on a nominal scale of two values with an error ∆=0,1. Feature weights: $w^1 = w^2 = 0,5$. Figure 11 shows the simulation results with the distribution of IOs in the space of their planar coordinates, where IOs from two different sources are shown as dots of different shapes. IOs for which the quantitative-qualitative proximity measure value exceeds 0.01 are encircled by a semi-transparent circle. For all IOs (except one pair, for which $\rho'_{Y_{ij}} = 0,29$ (additionally highlighted by a rectangle)), the object type values coincide. As the figure shows, the proximity measure value increases non-linearly as the linear distance between IOs decreases.

Furthermore, a mismatch in IO type values leads to a significant reduction in IO "similarity", despite their spatial proximity.

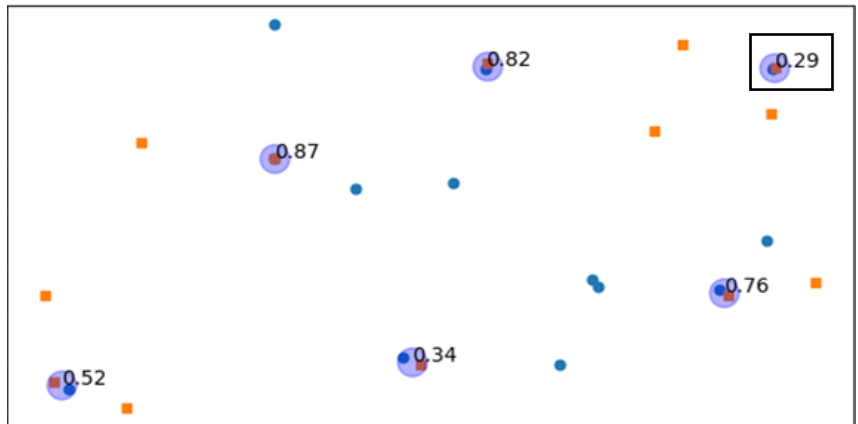

**Figure 11** – *Simulation results for generalized proximity measure (Sources: 20m/30m accuracy).*

Let us conduct an analogous experiment for the case of higher precision information sources (coordinate determination RMSE of 10 and 15 meters). The modeling results, with all other conditions remaining identical, are shown in Figure 12.

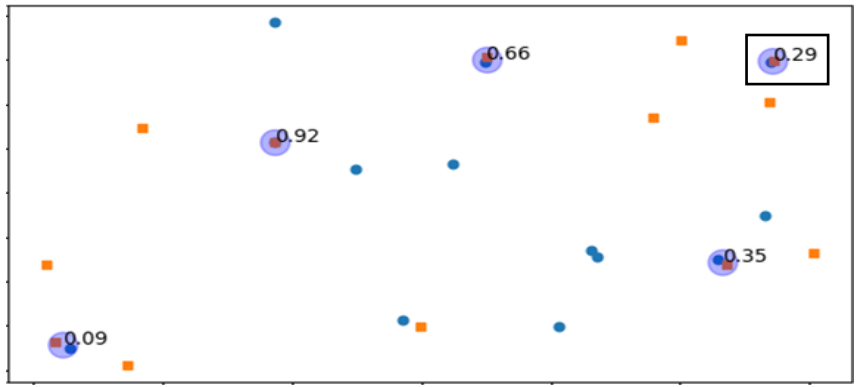

**Figure 12** – *Simulation results for generalized proximity measure (Sources: 10m/15m accuracy).*

Comparing the results shown in Figures 11 and 12, it may be noted that for objects in very close linear proximity, the value of the proposed proximity measure increased further due to the higher confidence (high precision) in the information sources. Conversely, for IOs separated by a greater linear distance, the proximity measure value decreased due to the lower possibility that such a distance could be obtained when using more precise sources. Regarding the two "linearly very close" objects of different types (marked by rectangles in both figures), the distance between them remained almost unchanged due to the multiplicative influence of the IO indicators on the measure value.

**Conclusions**

The proposed quantitative-qualitative proximity measure (or its inverse distance measure) is theoretically grounded and may be used as a universal measure to solve the problem of identifying information objects, data on which enter the information environment from various sources, by accounting for possible errors in their determination by such sources. This will allow to reduce the information load on system users (increasing the level of automation of the information processing process) and eliminate unwanted data duplication within the system. Such duplication leads not only to an unproductive increase in data volumes within the information resource but also to an erroneous assessment of the object saturation of the external environment analyzed by the system, and consequently, to an increased probability of erroneous decisions.

Unlike many known measures, the proposed measure does not require the transformation of feature values to ensure their comparability. The paper also proposes several measure variants for determining IO proximity based on all features. The obtained multiplicative proximity measure based on all features may be used to solve the problem of metric recognition of objects described by a set of features whose values are determined with errors. A limitation of the approach is the necessity of a priori specification of measurement errors and fuzzy set parameters.

To solve the IO identification task considered in this work, future research should focus on selecting an effective method for the automatic grouping of IO candidates for identification using the proposed measure.

**References**

[1] I.I. Obod, I.V. Svyd, and O.S. Maltsev, *Data Processing of Airspace Surveillance Radar Systems: Tutorial*. Kharkiv: Madryd, 2021, 255 p.

[2] V.Ye. Muhin, V.V. Zavgorodniy, Ya.I. Kornaga, and L.V. Baranovska, "Specialising algorithm for the information space formation," in *Information Technology and Security: Papers*

*of XXI International Scientific and Practical Conference (ITS-2021)*, vol. 21. Kyiv: Engineering, 2021, pp. 123–128.

[3] I.D. Mandel, *Cluster Analysis*. Moscow: Finansy i Statistika, 1988, 176 p.

[4] V. Hryhorovych, "Analysis of metrics for intelligent information systems," *Information Systems and Networks*, vol. 9, pp. 96–111, 2021. DOI: https://doi.org/10.23939/sisn2021.09.096.

[5] A.O. Fedorov, P.V. Notovskiy, and A.E.Yu. Peredriy, "Using the proximity measure in the problem of distributing the production program by planning periods based on the similarity coefficients of Dake, Jaccard, Maxfedor, Otiai, Rao, Tanimoto," *Bulletin of the National Technical University "KhPI". Economical Science*, no. 1 (3), pp. 32–35, 2020. DOI: https://doi.org/10.20998/2519-4461.2020.1.32.

[6] I.P. Gamayun and O.M. Bezmenova, "Formation of similarity indicators between objects characterised by parameters measured in different measurement scales," *Bulletin of the National Technical University "KhPI"*, no. 55 (1097), pp. 88–91, 2014. DOI: https://doi.org/10.20998/%x.

[7] "Measurement scales." [Online]. Available: https://elib.lntu.edu.ua/sites/default/files/elib_upload/%D0%95%D0%9D%D0%9F_%D0%AF%D0%BA%D0%B8%D0%BC%D1%87%D1%83%D0%BA_%D0%A1%D0%B5%D0%BB%D0%B5%D0%BF%D0%B8%D0%BD%D0%B0/page8.html. [Accessed: 07-Feb-2026].

[8] D. Dubois and H. Prade, "Possibility theory, probability theory and multiple-valued logics: A clarification," *Annals of Mathematics and Artificial Intelligence*, vol. 32, pp. 35–66, 2001.

[9] L.B. Levenchuk, O.L. Tymoshchuk, V.H. Guskova, and P.I. Bidyuk, "Uncertainties in data processing, forecasting and decision-making," *System Research & Information Technologies*, no. 3, pp. 66–80, 2023. DOI: 10.20535/SRIT.2308-8893.2023.3.05.